\documentclass[runningheads]{llncs}

\usepackage{eccv}

\usepackage{eccvabbrv}

\usepackage{graphicx, color}
\usepackage{booktabs}
\usepackage{soul}
\usepackage{subcaption}
\usepackage[utf8]{inputenc}
\usepackage{amsfonts}
\usepackage[english]{babel}
\usepackage{multirow}
\usepackage{enumitem}

\usepackage[accsupp]{axessibility}  

\usepackage{hyperref}

\usepackage{orcidlink}

\begin{document}

\title{Context-Infused Visual Grounding for Art} 


\author{Selina Khan\inst{1}\orcidlink{0000-0002-6443-8250} \and
Nanne van Noord\inst{1}\orcidlink{0000-0002-5145-3603}}

\authorrunning{Selina Khan and Nanne van Noord}

\institute{University of Amsterdam \\
\email{selinajasmin@gmail.com, n.j.e.vannoord@uva.nl}}

\maketitle

\begin{figure}[!ht]
    \vspace{-5mm}
    \centering
    \includegraphics[width=0.9\textwidth]{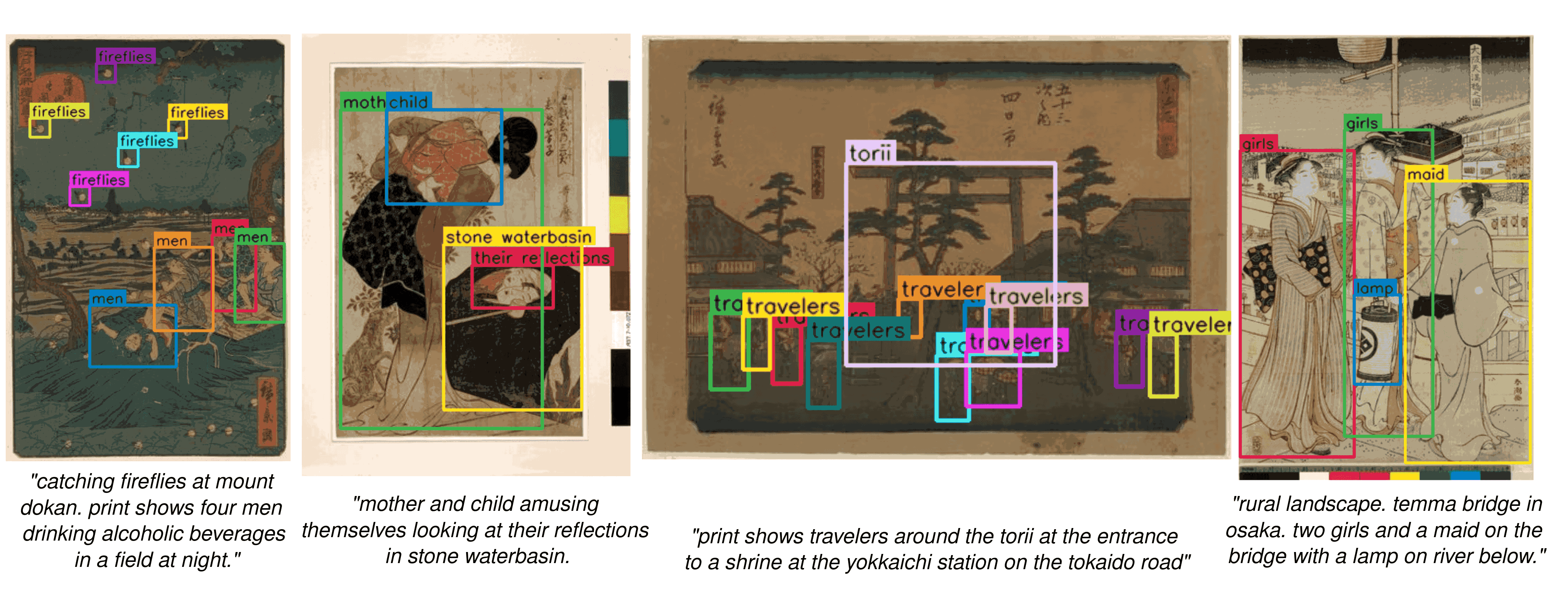}
    \caption{Example entries from the Ukiyo-eVG dataset.}
    \label{fig:ukiyo-ex}
    \vspace{-8mm}
\end{figure}

\begin{abstract}
Many artwork collections contain textual attributes that provide rich and contextualised descriptions of artworks. Visual grounding offers the potential for localising subjects within these descriptions on images, however, existing approaches are trained on natural images and generalise poorly to art. In this paper, we present CIGAr (Context-Infused GroundingDINO for Art), a visual grounding approach which utilises the artwork descriptions during training as context, thereby enabling visual grounding on art. In addition, we present a new dataset, Ukiyo-eVG, with manually annotated phrase-grounding annotations, and we set a new state-of-the-art for object detection on two artwork datasets.
  
\end{abstract}

\section{Introduction}
\label{sec:intro}
Imagery from our cultural heritage are a vital source of knowledge, providing insights into human history and helping us interpret our collective past. As such, preserving these collections is essential to protect our human cultural identity \cite{culher}. In efforts to do so, there has been a growth in the digitisation of such collections on a large scale \cite{wikiart, moma, omniart}. Art historians increasingly use these digital collections for their research, underscoring the importance of developing efficient tools to support the analysis of these large databases \cite{digcultures, artappl}.

Existing AI approaches for stylistic analysis of artwork images excel in learning visual features and linking them to various attributes  \cite{rijksmuseum, artofdetection, survey}. However, these methods often rely on annotated datasets, which are time-consuming to create and require domain-expert knowledge. Although techniques like transfer learning \cite{deepart, arttl, cecipipe, insearchofart} and weakly supervised learning \cite{wsod2018, wsod2022, wsodinoue} have shown promise to mitigate these challenges within the artistic domain, difficulties in generalising across domains persist \cite{transflimit, wsodlimit}.

Approaching digital artwork analysis in a task-specific manner requires delineating relevant attributes for specific tasks. This allows focused research on particular attributes but often fails to capture the nuanced historical context critical for in-depth art analysis. The historical context includes the period, cultural and social influences, and provides valuable insights into the artist's intentions and the impact of the artwork. Current digital art analysis primarily targets the artwork's objective content, utilising image-level annotations like origin, associated art movement for attribute classification \cite{ftartclas, omniart, wikiart}, or labelled objects from the depicted scene for object detection \cite{wsod2022, peopledet}. Relying solely on these objective properties can limit detailed analysis by neglecting broader contexts that are harder to capture in explicit categories. The individual portrayal and contextual setting play a crucial role in art.

Despite progress in incorporating context into digital art analysis through multi-task learning \cite{artcontext, ukiyoe, omniart}, the challenge of integrating broader artwork context without explicit annotations remains under-explored. Recognising the importance of broader contexts in analysing digital art underscores the need for models capable of understanding the cultural context of an artwork. This is essential to identify significant visual elements in various contexts. Capturing such nuanced attributes is challenging due to the difficulty in encapsulating them in explicit annotations.

To address this gap, we propose to leverage artwork titles and descriptions, which are rich in context, to enhance detection in artwork images. These often provide detailed explanations of the depicted scenes and highlight significant elements as defined by artists or experts. Our approach integrates these textual descriptions into the training pipeline to detect and classify entities within the artwork through Visual Grounding (VG). VG models align visual elements with textual queries, allowing for a more fine-grained understanding of the context of visual elements \cite{vg}. 

For this work, we build on Transformer-based VG models, which excel in zero-shot scenarios with natural images, highlighting their utility when limited ground-truth data is available \cite{groundingdino, glip}. Typically, such models are trained with an object detection objective on large-scale image-text datasets. For domain adaptation on (typically smaller) collections of artworks, however, we find that incorporating context during training is crucial to learning relevant features - as depictions of subjects may differ significantly across artworks. 

To this end, we propose the following contributions:
\begin{enumerate}[topsep=0pt]
    \item An approach for integrating context within visual grounding for art. 
    \item The first visual grounding dataset for artworks.
    \item State-of-the-art results on the IconArt and ArtDL artwork datasets. 
\end{enumerate}

\section{Related Work}
\label{sec:relwork}
Within this work we propose to approach artwork analysis from the perspective of Visual Grounding, in the following we give an overview of relevant work on Visual Grounding and discuss its relation with Object Detection, which has been more commonly applied to artworks.

\subsection{Visual Grounding}
Integrating visual and textual information through a joint vision-language embedding as core representation leads to better cross-task transfer and improvement in visual recognition \cite{alignedimword, imwordemb3}. Open-set object detection leverages these embeddings to identify arbitrary classes during inference through language generalisation \cite{osod}. Models like CLIP \cite{clip} and ALIGN \cite{align} use cross-modal contrastive learning on vast image-text pairs, enabling open-vocabulary image classification. ViLD \cite{vild} extends this by distilling knowledge from CLIP/ALIGN into a two-stage detector for enhanced object detection, while MDETR \cite{mdetr} uses a modified DETR loss for open-vocabulary training, excelling in long-tail object categories.
OV-DETR \cite{ovdetr} utilises CLIP for image-text embeddings, decoding category-specific boxes using the DETR framework, proposing the first end-to-end Transformer-based open-vocabulary detector. GLIP \cite{glip} reformulates object detection as a grounding problem, pre-training on both detection and grounding data to learn aligned semantics. DetCLIP \cite{detclip} enhances this with pseudo labels from large captioning datasets. Current state-of-the-art VG model GroundingDINO \cite{groundingdino} improves GLIP by extending the fusion of image and text features during training and evaluating the adaptability of the model to novel objects described with attributes in a Referring Expression Comprehension (REC) task.
Although including REC datasets extends the evaluation of open-set object detection to include linguistically complex inputs, their abilities are difficult to evaluate in current VG benchmarks \cite{refcoco, refcocog, referit, clevr} as they consist mainly of simple descriptive texts \cite{vgsg}. Efforts to include additional knowledge into the VG pipeline, like graphical scene representations \cite{graphknowledge, linggraph, objsg}, reasoning modules \cite{iterreason}, and human-annotated scene descriptions \cite{vgsg} show improved reasoning. 

This work incorporates artwork-specific context during training by leveraging titles and descriptions in the VG pipeline.

\subsection{Object Detection in Artworks}
Object detection in artistic images aids tasks like facial expression analysis \cite{facedet, facedet2, facedet3}, image captioning \cite{artcap}, and enhancing augmented reality in galleries \cite{arart}. Despite its importance, it remains challenging due to the diversity of object representation across art styles, the domain shift between natural images and artworks \cite{survey, artofdetection}, and the high cost of annotation. Object detection datasets have historically comprised only a minor fraction of the broader spectrum of art datasets, which can be partly attributed to these challenges. Nonetheless, there have been significant efforts in the development of object detection datasets applied across a variety of settings, adopting diverse training methods to accommodate the unique characteristics of artworks \cite{survey}. 

Significant results have been achieved using pre-trained models on datasets like Artistic Faces \cite{artfaces}, KaoKore \cite{kaokore}, and PeopleArt \cite{peopleart}, demonstrating good generalisation in semi-automatic annotation of faces \cite{kaokore2, faceofart}, transfer learning \cite{wsod2022, peopleart}, and style transfer settings \cite{styletransfer, oadataaug}. Datasets like Paintings \cite{paintingadata}, Watercolor2k, Clipart1k, and Comic2k \cite{wsodinoue} include PASCAL-VOC \cite{pascal} classes, while CASPApaintings \cite{caspa} covers animal COCO \cite{coco} categories, showing impressive object-detection performance but primarily for man-made objects and animals \cite{artofdetection, wsodinoue}. Motivated by the lack of object detection annotations and the specificity of categories of interest, several works have adopted a Weakly Supervised Object Detection (WSOD) learning strategy, utilising datasets with art-specific object attributes \cite{wsod2022, artdl}. 
To further enable research on object detection, the DEArt (Dataset of European Art) \cite{deart} was developed consisting primarily of art-specific annotations, in combination with standard COCO classes. Recently, image descriptions for nearly half of the images from the original DEArt dataset were made available \cite{deartdesc}, which we utilise for visual grounding. 
 
Despite their promise, we believe that the effectiveness of these models is still limited by the complex semantics of the dataset's classes. Challenges arise from overlapping semantics between classes, such as \texttt{`nudity'} and \texttt{`child Jesus'}, and annotation inconsistencies, such as images with multiple visually similar saints where only some are labelled. These issues negatively impact model performance, leading to mistakenly disregarding culturally significant objects due to annotation limitations.

To mitigate the limitations posed by a closed set of annotated objects, this work incorporates the artwork's titles and descriptions in the object detection pipeline, by reformulating it as an open-vocabulary phrase grounding task. 
Approaching digital artwork analysis as a vision-language task by incorporating descriptions as an additional modality has been primarily done in the context of image captioning \cite{descrmatch, capgen, matching,iconcap} and image retrieval \cite{semart}. To the best of our knowledge, this has not yet been done in the context of object detection.

\vspace{-2mm}
\section{Ukiyo-eVG}
\label{sec:datasets}
\vspace{-1mm}
The Ukiyo-e dataset as introduced in \cite{ukiyoe}, consists of 178.000+ images of Japanese woodblock prints, annotated with various artwork attributes and used for multi-task classification and regression tasks. Since the dataset contains rich titles and descriptions, we adopt a subset of the dataset to develop the first artwork dataset for visual grounding. The development of this dataset comprised three steps: (1) filtering based on subject-specific words in the titles and/or descriptions, (2) cleaning the titles and descriptions utilising a Large Language Model (LLM), and (3) generating pseudo-ground truth phrase grounding annotations.

\subsection{Data Collection}
\textbf{Filtering}. To create a phrase-grounding subset from the original Ukiyo-e dataset, we first select entries based on a closed set of keywords present in the titles and descriptions. These keywords are manually picked and include words that generally represent human subjects such as \texttt{`woman'}, \texttt{`courtesan'} and \texttt{`child'} within the domain of Japanese artworks (full list in the Appendix). Filtering images using this method has resulted in a dataset of 10993 images. We concatenate the title and description if at least one of them has a relevant keyword. 

\noindent \textbf{Cleaning}. To further refine the quality of the data to be suitable for phrase grounding, the descriptions are cleaned using GPT-4 \cite{gpt4} by prompting the LLM to remove all information not relevant to the content. A lot of the artwork descriptions include information about for example the auction house, physical condition and size of the print which are not meaningful in a phrase-grounding setting. The full prompt can be found in the Appendix.  

\noindent \textbf{Pseudo-ground truth}. Lastly, a pseudo-ground-truth is generated using the cleaned descriptions in zero-shot VG model GroundingDINO \cite{groundingdino} (GD). GroundingDINO is utilised as a baseline for all experiments in this research as it provides state-of-the-art results on various benchmarks. We run inference on GroundingDINO with the cleaned caption as a textual prompt alongside the artwork image using a text and bounding box confidence threshold of $0.20$ to obtain our initial pseudo-ground-truth proposals. Finally, the Ukiyo-eVG dataset is split into 80\% training, 10\% validation and 10\% test set.

\subsubsection{Annotation}
\label{sec:annot}
To obtain a reliable ground truth for evaluation, we have manually annotated the 1100 test set images. Initially, a baseline set of annotations has been generated through inference on a pre-trained GroundingDINO model, which we then manually refined. We have developed a simple but effective annotation tool\footnote{Annotation tool available at: \url{https://github.com/samtitar/box-annotator}} to load and correct these predefined annotations. This tool allows for free-form annotation, enabling the editing of individual phrases, boxes, and image captions.

We recognise the cultural significance of these carefully crafted descriptions and aim to annotate exactly what is described in the original text, avoiding personal interpretations. In two common ambiguous cases, we have made specific annotation decisions. Firstly, when the description is uncertain about the exact label to assign to an object, such as \texttt{`building or temple'}, we retain the ambiguity in the annotated phrase. Secondly, when multiple objects are referred to by multiple phrases and the visual appearance could not be directly linked to a specific phrase, such as \texttt{`travelers and porters'}, we split the annotation into multiple identical boxes, each referring to a different possible phrase. 

Annotated examples are shown in Figure \ref{fig:ukiyo-ex}, and the complete Ukiyo-eVG dataset is shared for re-use. \footnote{Ukiyo-eVG dataset available at: \url{https://ap.lc/VCUaA}}

\subsection{Pseudo-ground Truth Refinement}
\label{sec:pgt}
Initial qualitative results demonstrate the potential for zero-shot GroundingDINO as a source of pseudo-ground truth, yet, it tends to overpredict certain phrases due to uncertainty on the novel domain. To refine the pseudo-ground truth proposals for the Ukiyo-eVG dataset into more meaningful annotations, we develop a refinement algorithm which iteratively filters confident predictions on both phrase and box levels. More specifically, we refine the pseudo-ground truth in multiple stages, considering various properties of the box-phrase proposals at each stage. The first stage filters out individual proposals with generic phrases describing non-objects such as \texttt{`print'}, \texttt{`scene'} and \texttt{`image'}. 

\begin{figure}[!ht]
    \centering
    \includegraphics[width=\textwidth]{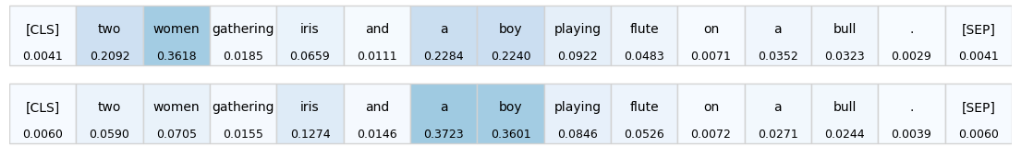}
    \caption{Text token distribution of the input prompt on two identical box proposals. Box 1 (top) predicts the multi-token phrase \texttt{`two women a boy'} as phrase groups \texttt{`two women'} and \texttt{`a boy'} both exceed the base threshold of $0.20$. Box 2 (bottom) shows a correct prediction referring to \texttt{`a boy'}.}
    \label{fig:textdist}
    \vspace{-4mm}
\end{figure}

In the second stage, we refine the model's prediction abilities on a per-box phrase level. Open-vocabulary VG models predict a distribution over the tokens in the input for each object and are optimised to equally spread the weights over all tokens referring to the predicted box. Because GroundingDINO is not optimised for text data that describe scenes in Japanese woodblock prints, the confidence of each proposal is relatively low. This causes the model to often predict multiple boxes for the same object, or predict a single box spanning multiple phrases. In Figure~\ref{fig:textdist}, this issue is illustrated with two phrase predictions for two identical box proposals. We utilise the text token predictions to first un-group multiple phrase predictions for individual boxes and reduce it to a single phrase prediction with the highest confidence. Then, identical boxes are filtered based on the highest average confidence between all tokens in the phrase. In the example from Figure~\ref{fig:textdist}, this results in box 2 with the phrase \texttt{`a boy'} being preferred over box 1 with the phrase \texttt{`a woman'} and is ultimately correct. Only if the phrases from both boxes are separated by ``\textit{of}'' or ``\textit{as}'' in the input prompt, we merge the phrase. This ensures that expressions in the form \texttt{[subject]~as~[role]} are not filtered out.

The last refinement stage attempts to mitigate noisy predictions based on the localization abilities of the model. We first post-process all phrases by removing quantity descriptors and articles to ensure consistency between phrases referring to the same object. Then, a final refinement step filters boxes positioned within other boxes with the same phrase based on confidence scores.      

\subsubsection{Application of Refinement Algorithm}
We refine the pseudo-ground-truth for all data splits which do not contain ground truth annotations. For visual grounding, we refine the pseudo-ground truth of the Ukiyo-eVG train and validation splits, and for the entire DEArt Captions \cite{deartdesc} dataset.  For object detection, we use the ArtDL \cite{artdl2} and IconArt \cite{wsod2018} datasets, which include a labelled test set for object detection, but only contain class labels (without boxes) for the training set. Consequently, we generate and refine the pseudo-ground truth for the class-labelled data through a text prompt comprised of all present classes in the artwork image. Lastly, the DEArt object detection dataset \cite{deart} includes ground-truth annotations and thus does not need pseudo-ground truth.

When training a model that performs text-image matching, we generally expect minor deviations when matching ground truths to model predictions. The phrase \texttt{`child'} and \texttt{`a child'} might refer to the same thing but would be misclassified when requiring an exact phrase match. A solution to this can be fuzzy phrase matching, which classifies phrases as correct if they overlap by a certain threshold. When training on ground-truth data, we can allow this degree of freedom in the phrase-matching process, as there is a likely guarantee that the fuzzy matching will match the correct label. When training on pseudo-ground-truth data, however, allowing for fuzzy phrase matching might introduce too much noise to the optimisation process as we cannot guarantee that the matched phrase is fully correct. Moreover, to compare the performance of a model trained on ground-truth data we evaluate performance on fuzzy-matched ground-truth data with exactly matched refined pseudo-ground-truth data. For each dataset with annotated test data, we compare visual grounding and object detection inference results adopting exact phrase matching with refined data, and fuzzy matching with ground-truth data. 

When observing the results in Table \ref{tab:phrasematching}, we notice that fuzzy alignment on raw inference data and exact matching on a refined dataset are very close in performance, suggesting that the refined pseudo-ground truth labels are a good alternative to annotated ground truth. In particular, for the Ukiyo-eVG and ArtDL datasets, we notice only a minor decrease ($\approx 11\%$) after refinement compared to the exact ground truth. The IconArt dataset drops a bit more in trainable performance, which we suspect is due to the large variance in specificity between objects in the labels. Nonetheless, when comparing the baseline performance on exact matching with all refined datasets, we observe a significant increase in trainable model performance, confirming that our refinement algorithm is an effective method for pseudo-ground-truth generation.    
 
\begin{table}[!ht]
\centering
\begin{tabular}{l|c|ccc}
\hline
\multicolumn{1}{c|}{\textbf{Dataset}} & \multicolumn{1}{c|}{\textbf{Phrase Matching}} & \textbf{mAP} & \textbf{mAP@0.5} & \textbf{R@1} \\ \hline \hline
\textbf{Ukiyo-eVG}           & fuzzy & \textbf{54.3} & \textbf{58.5} & \textbf{45.5} \\
\textbf{Ukiyo-eVG}           & exact  & 19.7         & 20.9         & 17.5         \\
\textbf{Ukiyo-eVG - refined} & exact & 48.1          & 52.5          & 42.2          \\ \hline
\textbf{ArtDL}               & fuzzy & \textbf{47.4} & \textbf{64.6} & \textbf{55.6} \\
\textbf{ArtDL}               & exact & 18.7          & 30.4          & 20.7          \\
\textbf{ArtDL - refined}     & exact & 42.0          & 62.1          & 56            \\ \hline
\textbf{IconArt}             & fuzzy & \textbf{18.7} & \textbf{30.4} & \textbf{20.7} \\
\textbf{IconArt}             & exact & 5.4           & 9.2           & 20.9          \\
\textbf{IconArt - refined}   & exact & 12.5          & 20.6          & 20.7          \\ \hline
\end{tabular}
\vspace{2mm}
\caption{Effectiveness of pseudo-ground-truth refinement on datasets without ground truth training annotations. All results on annotated ground truth test sets.}
\label{tab:phrasematching}
\vspace{-8mm}
\end{table}

\subsection{Dataset Statistics}
\label{sec:stats}
Statistics for all datasets used in this research along with their (refined pseudo-) ground-truth data splits are shown in Table \ref{tab:datasets}. The Ukiyo-eVG test split contains 3880 box-phrase annotations (1482 unique phrases). The DEArt Captions subset includes 4 or 5 human-annotated captions per image for 7471 images, which comprises nearly half of the images from the DEArt dataset. Our experiments on visual grounding therefore adopt a different split as used in the paper by the original authors \cite{deart}, as this does not include captions. We select the subset of the original DEArt dataset that includes captions and expand this into the DEArt Captions set to include one data entry per unique caption. This results in 4/5 identical image entries in the dataset, each with a different caption but identical object detection annotations. This data set was divided into 80\% for training, 10\% for validation, and 10\% for testing and ensuring no overlap between identical images per set. This set is used for both visual grounding and object detection experiments. Additionally, we use the original DEArt dataset for experiments on object detection and use the same train-val-test split of 70/15/15 presented in \cite{deart}. Although the ArtDL dataset contains both saints and symbol annotations, we adopt the same label sets used in prior work \cite{artdl2}, which provides results for the 10 classes of Christian saints. Previous work on the IconArt dataset reports results on the subset of annotated test data, which includes both localisation and classification annotations for $7$ classes portraying religious subjects and objects \cite{artdl2, wsod2018, wsod2022}, which we use to evaluate our main experiments. Three additional classes (\texttt{`beard'}, \texttt{`turban'} and \texttt{`capital'}) are included in the dataset but remain unused in the present literature. We also perform experiments including these additional three classes.

\begin{table}[!ht]
\centering
\begin{tabular}{l|l|ccc|c}
\hline
\multicolumn{1}{c|}{\multirow{2}{*}{\textbf{Task}}} &
  \multicolumn{1}{c|}{\multirow{2}{*}{\textbf{Dataset}}} &
  \multicolumn{3}{c|}{\textbf{Ground Truth}} &
  \multirow{2}{*}{\textbf{\# Images}} \\ \cline{3-5}
\multicolumn{1}{c|}{}                      & \multicolumn{1}{c|}{} & \textbf{Train} & \textbf{Val} & \textbf{Test} &             \\ \hline \hline
\multirow{2}{*}{\textbf{Visual Grounding}} & Ukiyo-eVG             & \texttimes              & \texttimes             & $\checkmark$             & $\sim$11000 \\
                                           & DEArt Captions \cite{deartdesc}       & \texttimes               & \texttimes             & \texttimes              & $\sim$33000 \\ \hline
\multirow{4}{*}{\textbf{Object Detection}} & DEArt \cite{deart}                & $\checkmark$               & $\checkmark$             & $\checkmark$              & $\sim$15000 \\
                                           & DEArt Captions \cite{deartdesc}        & $\checkmark$               & $\checkmark$             & $\checkmark$              & $\sim$33000 \\
                                           & ArtDL \cite{artdl2}                 & \texttimes*               & $\checkmark$             & $\checkmark$              & $\sim$24000 \\
                                           & IconArt \cite{wsod2018}               & \texttimes*               & -            & $\checkmark$              & $\sim$5000  \\ \hline
\end{tabular}
\vspace{2mm}
\caption{Overview of all datasets used in this research. For datasets with no ground truth available, we generate refined pseudo-ground-truth data to suit the task. Datasets marked with * include per-image class labels.}
\label{tab:datasets}
\vspace{-13mm}
\end{table}

\section{CIGAr}
\label{sec:cigar}
Transformer-based VG models use context-rich embeddings to align text and image features, grounding phrases from descriptive prompts. Models like GLIP \cite{glip} and GroundingDINO \cite{groundingdino} treat phrase grounding as object detection, utilising large image-text databases to generalise across various appearances and expressions. However, for domain adaptation to artworks, context becomes crucial when training as the depiction of relevant subjects may differ significantly across artworks and within certain contexts. This work addresses this by incorporating scene descriptions during training to provide the necessary context.

\begin{figure}[!ht]
    \centering
    \includegraphics[width=0.85\textwidth]{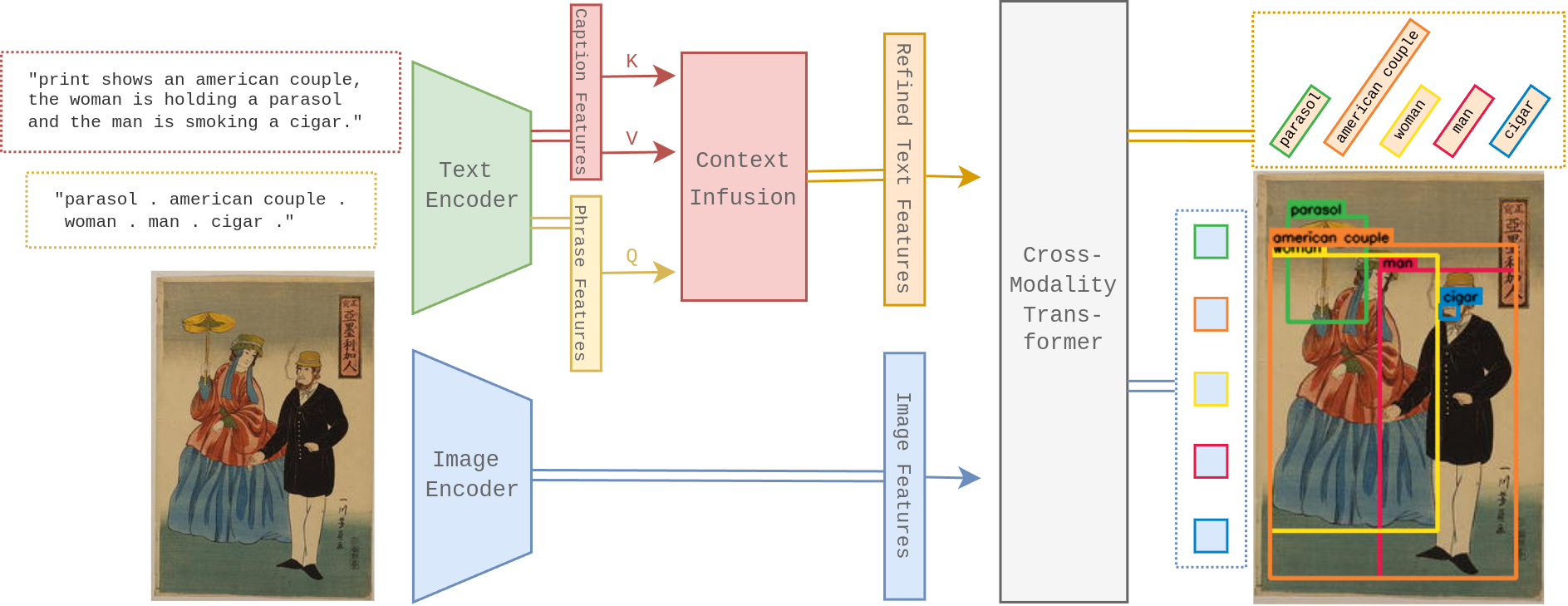}
    \vspace{-1mm}
    \caption{Overview of CIGAr. Components marked in red are additions to the GroundingDINO model architecture. Phrase and caption embeddings are extracted from a fine-tuned BERT encoder after which they are fused to provide context-rich phrase embeddings. They are passed through a cross-modality transformer alongside the image features before matching the predicted phrases and object boxes.}
    \label{fig:architecture}
    \vspace{-5mm}
\end{figure}

In turn, we propose CIGAr (\textbf{C}ontext-\textbf{I}nfused \textbf{G}roundingDINO for \textbf{Ar}t), an extension of GroundingDINO that incorporates scene context during training. An overview of CIGAr is shown in Figure \ref{fig:architecture}. We extract text embeddings for both the artwork description and grounding phrases, which are merged in a learnable cross-attention mechanism. By passing the artwork description as keys and values, and the phrases as queries, the model effectively learns what the phrases mean in the context of the depicted scene before aligning with the image regions. The refined context-infused text features are consequently fused with image features in the cross-modality transformer before making box-phrase alignment predictions. By introducing context in the phrase embedding, the model can perform the detection in a phrase-box alignment manner, without needing excessive adaptations to the training mechanism. 

Besides context infusion in the phrase embedding, we introduce domain-specific knowledge in the model by fine-tuning the text backbone on artwork-specific entities. Experiments on the effectiveness of this approach are highlighted in Section \ref{sec:ftbert}. 

\subsection{Text Encoder Fine-tuning} 
\label{sec:ftbert}
To align text and image features, the VG model needs to learn which phrases in a natural language query are relevant objects to be localised in the image. This ability is largely attributed to the natural language understanding capabilities of the text encoder \cite{glip}, which is particularly challenging when dealing with a domain shift to artwork descriptions. For this reason, we enrich the model's domain knowledge on relevant entities by fine-tuning a pre-trained BERT \cite{bert} language model. 

Domain-related knowledge is introduced to a pre-trained BERT model by including object- and relation-aware scene knowledge through unlocalised scene graphs extracted from the artwork descriptions using the (Stanford) Scene Graph Parser (SGP) \cite{sgp} (See the appendix for an example). The SGP distinguishes entities and their relations in a purely rule-based manner, and therefore does not require prior domain knowledge. The parsed entities and relation triplets are used as pseudo-ground truths on which we evaluate the experiments.

We fine-tuned a pre-trained BERT model using masked language modelling and thoroughly experimented with various masking strategies for randomly masking entity- or relation-specific words during training as extracted from the scene graphs. For our experiments, we mask all present entities with their spans 90\% of the time, as this demonstrated the most significant improvement over the standard BERT model. We elaborate on the different masking strategies tested in the Appendix. 

To evaluate the learned embeddings, we use the fine-tuned BERT model as a backbone to perform Named Entity Recognition (NER). The model has to predict the span of all entities (and their modifiers) and all relation words used to describe any entity in relation to another from the description. The model consists of a single trainable linear layer on top of the (fine-tuned) BERT model. To fully exploit the available data from the Ukiyo-eVG dataset we use the train split to fine-tune the BERT model and re-split the concatenated validation and test set into a $80/20$ split for training and testing the NER task. For the DEArt Captions dataset, we use our original pre-defined splits. The train set is used to fine-tune the BERT model, the validation set to train the model to perform NER, and the test set to evaluate NER performance. 

\vspace{-5mm}
\subsubsection{Results}
\label{sec:ftbertresults}  
We compare the performance on the NER task between a pre-trained and our fine-tuned BERT model. The results are shown in Table \ref{tab:bertft}. The improved performance across all metrics indicates that fine-tuning the text encoder is beneficial for this task, and aids the model's understanding of the novel domain.

\begin{table}[!ht]
\vspace{-2mm}
\centering
\footnotesize
\begin{tabular}{l|c|cccc}
\hline
Dataset & \begin{tabular}[c]{@{}c@{}}\textbf{Fine-tuned} \\ \textbf{BERT}\end{tabular} & \textbf{Precision} & \textbf{Recall} & \textbf{F1} & \begin{tabular}[c]{@{}c@{}}\textbf{Per-entity} \\ \textbf{Accuracy}\end{tabular} \\ \hline \hline

\multirow{2}{*}{\textbf{Ukiyo-eVG}}  
& $\times$  & 70.8  & 76.0  & 73.1  & 59.6  \\
& \checkmark  & \textbf{73.8}  & \textbf{76.6}  & \textbf{75.1}  & \textbf{74.1}  \\ \hline

\multirow{2}{*}{\textbf{DEArt Captions} \cite{deartdesc}}  
& $\times$  & 82.3  & 88.1  & 85.6  & 84.2   \\
& \checkmark  & \textbf{86.2}  & \textbf{88.5}  & \textbf{87.1}  & \textbf{91.1}  \\ \hline

\end{tabular}
\vspace{2mm}
\caption{Performance of pre-trained and fine-tuned BERT on NER per-token (Precision, Recall and F1) and per-entity (Accuracy) on Ukiyo-eVG and DEArt Captions.}
\label{tab:bertft}
\vspace{-7mm}
\end{table}

In addition to measuring the performance based on exact token matches, we evaluate accuracy per whole entity as it assesses how well the model captures the full span of each entity. This evaluation is not performed for the relation words, as analysis showed that the fine-tuned BERT model does not significantly improve in recognising this word type. This can be attributed to the fact that relation-indicating words are not necessarily art-specific. The per-entity accuracy further highlights the ability of fine-tuned BERT to distinguish relevant entities from non-entities. 

\section{Experiments}
All experiments on both visual grounding and object detection employ the pre-trained GroundingDINO-T checkpoint\footnote{From \url{https://github.com/IDEA-Research/GroundingDINO}} that was trained on a vast collection of image-text pairs. Experiments on GroundingDINO-L concluded that because this model is trained on phrase grounding data, it proved difficult to fine-tune on domain-specific datasets.  All model hyper-parameters initially remain unchanged from the provided configuration associated with the model checkpoint \footnote{Training and configuration details on  \url{https://github.com/selinakhan/CIGAr}}. For all experiments, we freeze the text and image backbone and use a decreased learning rate of $1e-5$ for more stable fine-tuning and a batch size of $8$ on a single NVIDIA A100 GPU. Additionally, all experiments adopt early stopping regularisation.

\subsection{Visual Grounding}
Our main experiments assess the influence of including context and domain knowledge in the grounding pipeline for artworks. We evaluate the performance of CIGAr on two visual grounding datasets Ukiyo-eVG and DEArt Captions, and additionally carry out an ablation study on the added model components.     
\vspace{-3mm}
\subsubsection{Results}
\label{sec:vgresults}
For tasks like visual grounding with no consistent global class labels, it's generally more meaningful to calculate mAP on a per-image basis and then average these scores. This method respects individual contexts and variations in phrase-label mappings across images. All visual grounding experiments are evaluated in this manner.
 
\begin{table}[!ht]
\footnotesize
\centering
\begin{tabular}{l|l|ccccc}
\hline
\textbf{Dataset}                    & \textbf{Model}         & \textbf{mAP}  & \textbf{mAP@0.5} & \textbf{R@1}  & \textbf{R@10} \\ \hline \hline
\multirow{3}{*}{\textbf{Ukiyo-eVG }} & GD (Zero-Shot)                      & 19.7          & 20.9             & 17.5          & 20.1          \\
                                    & GD (Fine-tuned)                 & 18.4          & 21.1             & 16.6          & 18.6          \\
                                    & CIGAr                          & \textbf{25.1} & \textbf{26.8}      & \textbf{21.2} & \textbf{25.6}   \\ \hline
\multirow{3}{*}{\textbf{DEArt Captions}\cite{deartdesc} }     & GD (Zero-Shot)                      & 19.6          & 19.7             & 17.7          & 19.6          \\
                                    & GD (Fine-tuned)                            & 13.1          & 13.5             & 11.2          & 13.1          \\
                                    & CIGAr & \textbf{29.1} & \textbf{30.3}    & \textbf{25.8} & \textbf{29.5} \\ \hline
\end{tabular}
\vspace{2mm}
\caption{Grounding results of CIGAr, zero-shot and fine-tuned GroundingDINO model on Ukiyo-eVG and DEArt Captions test set. Best-performing model for each dataset is highlighted in bold.}
\label{tab:vgresults}
\vspace{-6mm}
\end{table}

Table \ref{tab:vgresults} summarises the results of our methods examined on Ukiyo-eVG and DEArt Captions. The performance of CIGAr is compared against a pre-trained and fine-tuned GroundingDINO model. We observe that CIGAr outperforms both the zero-shot and fine-tuned GD models for both visual grounding datasets, suggesting that our refined pseudo-ground-truth effectively utilises the provided context in the CIGAr model. A more significant increase in performance on the DEArt Captions dataset compared to Ukiyo-eVG is shown, which can potentially be attributed to the increased dataset size of DEArt Captions. On top of that, the DEArt captions were developed specifically to describe the content, whereas the Ukiyo-e artwork descriptions are generally written more lyrically, making them inherently more challenging to ground. Additionally, the Ukiyo-eVG dataset is evaluated on manually annotated data while DEArt Captions is evaluated on its refined pseudo-ground-truth. Interestingly, we observe a decrease in performance for both datasets after fine-tuning compared to zero-shot, highlighting the challenging nature of fine-tuning large pre-trained transformer-based models on small domain-specific datasets.

\begin{figure}[!ht]
\vspace{-5mm}
\centering
\begin{subfigure}[t]{.48\textwidth}
  \centering
  \includegraphics[width=\linewidth]{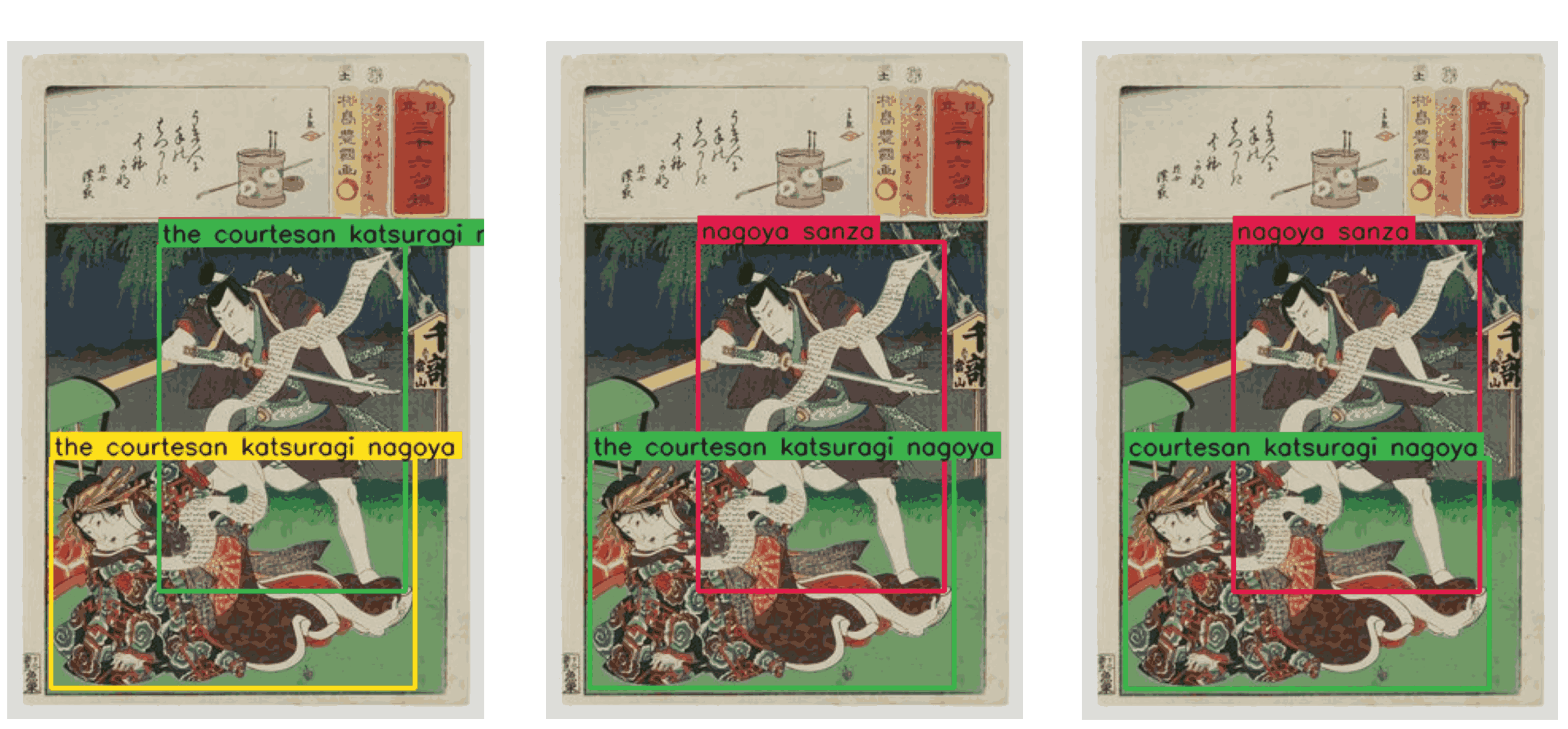}
  \caption{"nagoya sanza and the courtesan katsuragi nagoya"}
  \label{fig:nagoya}
\end{subfigure}%
\hspace{2mm}
\begin{subfigure}[t]{.48\textwidth}
  \centering
  \includegraphics[width=\linewidth]{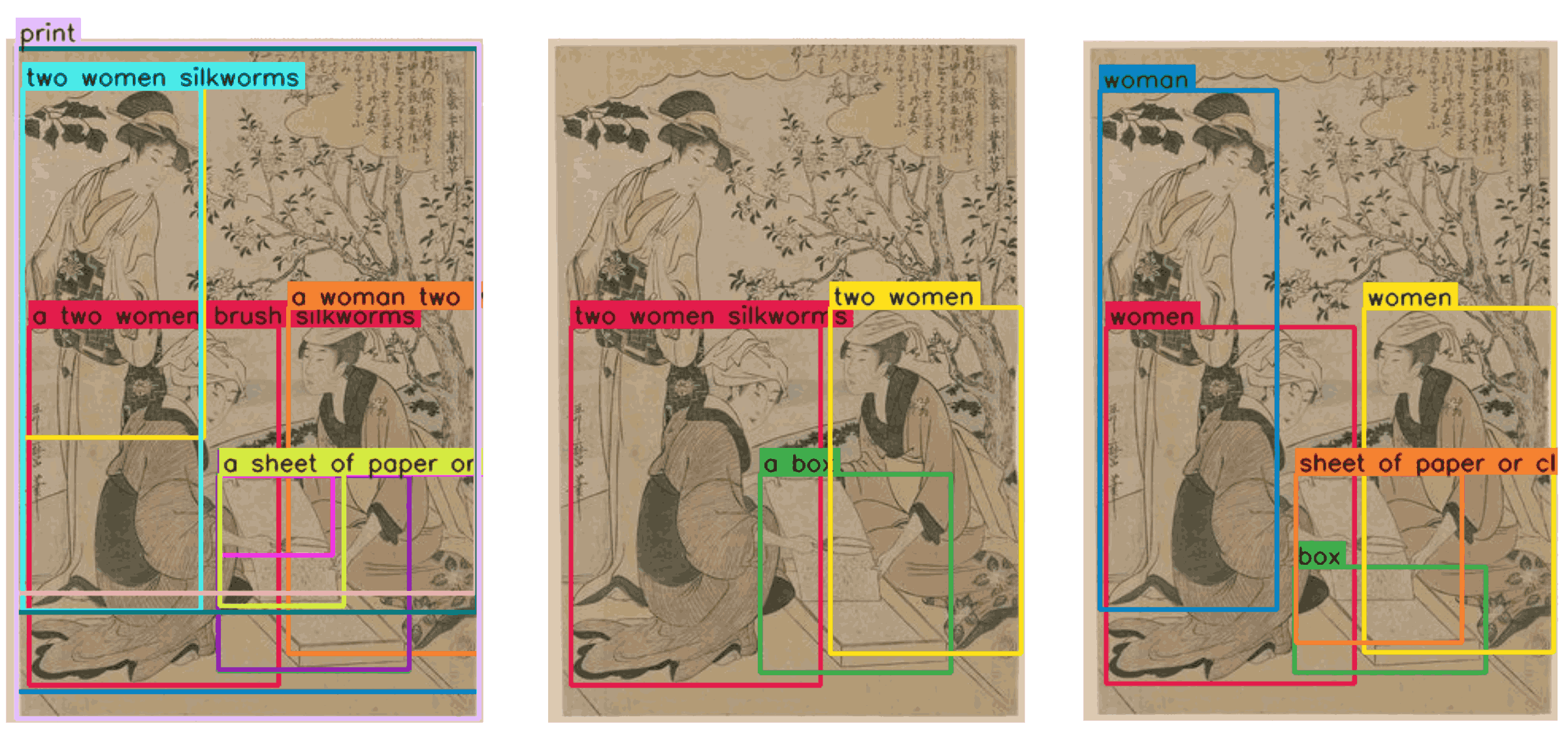}
  \caption{"print shows a woman watching as two women brush silkworms off a sheet of paper or cloth into a box."}
  \label{fig:silkworms}
\end{subfigure}
\vspace{-2mm}
\caption{Examples of CIGAr predictions on Ukiyo-eVG data comparing the zero-shot GD output (left) with the CIGAr output (middle) and ground-truth (right).}
\label{fig:qualuk}
\end{figure}
\vspace{-6mm}
Figure \ref{fig:qualuk} highlights two examples of the Ukiyo-eVG dataset before and after training. We notice that the model can distinguish which phrase in the caption corresponds to which present entity in the image in example \ref{fig:nagoya} after training. However, in our evaluation the phrase \texttt{`the courtesan katsuragi nagoya'} is considered incorrect even though semantically it is correct. This example emphasises the challenges of evaluating VG models as discussed before in \cite{mdetr}. As highlighted in \ref{fig:silkworms} we observe that CIGAr greatly decreases the rate of false-positive predictions and while it fails to predict objects it could in a zero-shot GD setting, it can correctly identify the \textit{`two women'} as referred to in the caption to be the two women in front.

\begin{table}[!ht]
\centering
\begin{tabular}{l|l|cccc}
\hline
\multicolumn{1}{c|}{\textbf{Dataset}} & \multicolumn{1}{c|}{\textbf{Model Ablations}} & \textbf{mAP} & \textbf{mAP@0.5} & \textbf{R@1} & \textbf{R@10} \\ \hline \hline
\multirow{3}{*}{\textbf{Ukiyo-eVG }}      
                                         & No context & 22.1          & 22.6          & 19.8          & 19.8          \\
                                         & Base BERT & 23.7          & 25.3          & 20.8          & 19.9          \\ 
                                         & CIGAr              & \textbf{25.1} & \textbf{26.8}      & \textbf{21.2} & \textbf{25.6} \\\hline
\multirow{3}{*}{\textbf{DEArt Captions}\cite{deartdesc} } 
                                         & No context & 19.2          & 19.7          & 17.8          & 19.3          \\
                                         & Base BERT & 27.2          & 28.0            & 24.5          & 27.3          \\ 
                                         & CIGAr              & \textbf{29.1} & \textbf{30.3} & \textbf{25.8} & \textbf{29.5} \\\hline
\end{tabular}
\vspace{2mm}
\caption{Ablations excluding the context infusion and the fine-tuned text backbone for CIGAr per dataset. }
\label{tab:ablations}
\vspace{-5mm}
\end{table}

The results in table \ref{tab:ablations} highlight our ablation experiments which exclude the fine-tuned BERT model (``Base BERT'') and the context-infusing mechanism (``No context'') from CIGAr. For both datasets, we observe increased results over the zero-shot GD baseline (from table \ref{tab:vgresults}). In particular, when combining both mechanisms in CIGAr performance remains best, suggesting that both components are needed to effectively utilise the provided context in this setting.

\subsection{Approaching Object Detection as Visual Grounding}
We further evaluate the potential of approaching closed-set object detection as phrase grounding by fine-tuning GroundingDINO on three established artwork datasets. We use our refined pseudo-ground-truth for training on ArtDL and IconArt and the provided ground-truth for DEArt and DEArt Captions. For the IconArt dataset, we train the model on the 10 present classes in the dataset and evaluate the 7 classes used in the prior literature for a fair comparison. Additionally, the infusion of context during training in a closed-set object detection setting is assessed using the DEArt Captions dataset, as it includes both object detection annotations along scene descriptions. Training a VG model on object detection data involves using all dataset classes as text input (dot separated) during both training and evaluation.

\vspace{-3mm}
\subsubsection{Results}
\label{sec:odresults}
From the results in table \ref{sec:odresults} we observe a significant increase in performance when training GroundingDINO on object detection data. In particular, utilising a transformer-based grounding model for pseudo-ground-truth extraction appears to improve the general object detection performance on the IconArt and ArtDL dataset by a significant margin. Performance on DEArt falls
behind, which can be due to the increased number of classes present compared to IconArt and ArtDL. Interestingly, GroundingDINO performance when trained on pseudo-ground-truth data seems to improve on prior weakly supervised benchmarks, while the model trained on DEArt, containing ground-truth data, does not. This suggests that this method provides a fruitful alternative to weakly supervised object detection methods. Experiments with CIGAr on DEArt Captions slightly reduce model performance, highlighting the difficulty of combining fixed object sets with open-ended scene knowledge. However, fine-tuning DEArt Captions improves performance, surpassing state-of-the-art results. Despite fewer unique entries, DEArt Captions outperforms the original dataset, suggesting that fine-tuning with duplicates acts as effective data augmentation.
 
\begin{table}[!ht]
\centering
\begin{tabular}{l|l|cccc}
\hline
\textbf{Dataset} & \textbf{Model}           & \textbf{mAP}  & \textbf{mAP@0.5}       & \textbf{R@1}  & \textbf{R@10} \\ \hline \hline
\multirow{3}{*}{\textbf{DEArt} \cite{deart}} & SOTA \cite{deart}                             & - & \textbf{31.2}          & - & - \\
        & GD (Zero-Shot)                   & 0.1  & 0.4           & 1.2 & 2.2 \\
        & GD (Fine-tuned)                  & 14.4 & 25.3 & 20.1 & 40.2 \\ \hline
\multirow{3}{*}{\textbf{DEArt Captions}\cite{deartdesc}}     & GD (Zero-Shot)                  & 6.1  & 9.5           & 14.2 & 28.7 \\
        & GD (Fine-tuned)                  & 35.1 & \textbf{50.1} & 38.3 & 61.2 \\
        & CIGAr & 25.5  & 37.5           & 34.1  & 58.5 \\ \hline
\multirow{3}{*}{\textbf{IconArt}\cite{wsod2018}} & SOTA \cite{artdl2}                             & - & 16.6          & - & - \\
        & GD (Zero-Shot)                  & 5.4  & 9.2           & 18.9 & 47.4 \\
        & GD (Fine-tuned)                  & 17.8 & \textbf{35.2} & 28.2 & 52.9 \\ \hline
\multirow{3}{*}{\textbf{ArtDL}\cite{artdl2}} & SOTA \cite{artdl2}                           & - & 41.5 & - & - \\
        & GD (Zero-Shot)                   & 1.5   & 2.1           & 25.6 & 66.8 \\
        & GD (Fine-tuned)                  & 35.6  & \textbf{51.1}           & 62.3 & 83.4 \\ \hline
\end{tabular}
\vspace{2mm}
\caption{Object detection results of zero-shot and fine-tuned GroundingDINO model on the IconArt, ArtDL and DEArt test set, and results of CIGAr on context-infused object detection on DEArt Captions. The best score per dataset is highlighted in bold.}
\label{tab:odresults}
\vspace{-5mm}
\end{table}

\begin{table}[!ht]
\centering
\resizebox{\textwidth}{!}{%
\begin{tabular}{l|cccccccccc|c}
\hline 
\multicolumn{1}{c|}{} & \begin{tabular}[c]{@{}c@{}}\textbf{Saint}\\ \textbf{Sebastian}\end{tabular} & \textbf{Turban} & \begin{tabular}[c]{@{}c@{}}\textbf{Cruxifiction}\\ \textbf{of Jesus}\end{tabular} & \textbf{Angel} & \textbf{Capital} & \textbf{Mary} & \textbf{Beard} & \begin{tabular}[c]{@{}c@{}} \textbf{Child}\\ \textbf{Jesus}\end{tabular} & \textbf{Nudity} & \textbf{Ruins} & \textbf{mAP@0.5} \\ \hline \hline
\textbf{SOTA} \cite{artdl2} & 22.1 & - & 58.9 & 0.9 & - & 1.9 & - & 1.7 & 24.3 & 6.1 & 16.6 \\
\textbf{CIGAr} & 2.8 & 25.0 & 40.4 & 14.6 & 0.00 & 34.7 & 47.3 & 82.9 & 30.9 & 40.2 & 31.8 \\ \hline
\end{tabular}%
}
\vspace{2mm}
\caption{Per-class AP on the IconArt dataset and mAP across all 10 classes.}
\label{tab:per_class}
\vspace{-8mm}
\end{table}

Finally, Table \ref{tab:per_class} highlights the per-class IconArt results on all 10 labels present in the original dataset. The class performance on the excluded labels \texttt{`beard'} and \texttt{`turban'} is particularly high, which can be attributed to the distinctive visual appearance of both objects, resulting in a reliable pseudo-ground-truth extracted in a zero-shot setting. Examples are provided in the Appendix. 

\section{Ethical Considerations}
This work aims to enable richer analysis based on captions associated with artworks. In doing so, we acknowledge that many art collections are built on a colonial heritage \cite{vawda2019museums} and that whilst these organisations are working to overcome issues stemming from this, it is still a common occurrence that the language used in collections is contentious (i.e., outdated or even offensive) \cite{brate2021capturing}. 
Methods, such as those proposed in this work, may inadvertently re-use contentious terms and reproduce the harms associated with them. Although for this work we did not encounter this, we would nevertheless emphasise that there is a \textit{duty of care} in applying vision-language models within such a historicised context.

\section{Conclusion}
This research introduces CIGAr (\textbf{C}ontext-\textbf{I}nfused \textbf{G}roundingDINO for \textbf{Ar}t), a visual grounding approach that uses artwork descriptions during training as context. We developed the Ukiyo-eVG dataset, the first visual grounding dataset for artworks, including a manually annotated test set. Our approach leverages pseudo-ground-truth data generated in a zero-shot setting, and algorithmically refines this for more reliable training annotations. Our approach significantly improves visual grounding performance on DEArt Captions and Ukiyo-eVG datasets. Additionally, fine-tuning GroundingDINO with our refined annotations achieved state-of-the-art results on IconArt and ArtDL, highlighting the effectiveness of approaching object detection in artworks through visual grounding.


%
%
\bibliographystyle{splncs04}
\bibliography{main}
\end{document}